%% file: main.tex
\documentclass{article}

\usepackage[final]{neurips_2024}

\usepackage{amsmath,amsfonts}
\usepackage{amssymb}
\usepackage{latexsym}
\usepackage{array}
\usepackage{textcomp}
\usepackage{url}
\usepackage{graphicx}
\usepackage{subcaption}

\usepackage{booktabs}
\usepackage{makecell}
\usepackage{paralist}
\usepackage{multirow}
\usepackage{multicol}
\usepackage{float}
\usepackage{tabularx}
\usepackage{longtable}
\usepackage{supertabular}
\usepackage{placeins} 
\usepackage{soul}     
\usepackage{xcolor}
\usepackage{hyperref}
\usepackage{orcidlink}
\usepackage{pifont}

\newcommand{\cmark}{\ding{51}}
\newcommand{\xmark}{\ding{55}}
\newcommand{\changes}[1]{\textcolor{black}{#1}}

\newcommand{\affUtor}{1}
\newcommand{\affUwm}{2}
\newcommand{\affVt}{3}
\newcommand{\affLoop}{4}
\newcommand{\affCmu}{5}
\newcommand{\affMiro}{6}
\newcommand{\affNvidia}{7}
\newcommand{\affPurdue}{8}
\newcommand{\affBuet}{9}

\usepackage{ragged2e}
\newcolumntype{P}[1]{>{\RaggedRight\arraybackslash}p{#1}}

\usepackage{enumitem}
\usepackage[T1]{fontenc}

\newcommand\BibTeX{{\rmfamily B\kern-.05em \textsc{i\kern-.025em b}\kern-.08em
T\kern-.1667em\lower.7ex\hbox{E}\kern-.125emX}}

\title{Mobile Imaging Solutions for Medical Diagnosis: Trends and Applications}

\author{%
  Syed Muhammad Ibne Zulfiker\textsuperscript{\affUtor} \\
  \texttt{syed.ibnezulfiker@utoronto.ca} \\
  \And
  Fariha Tabassum Islam\textsuperscript{\affUwm} \\
  \texttt{fislam2@wisc.edu} \\
  \AND
  Md Sultanul Arifin\textsuperscript{\affVt}\thanks{Corresponding author.} \\
  \texttt{sultanularifin@vt.edu} \\
  \And
  Khandker Aftarul Islam\textsuperscript{\affLoop,a} \\
  \texttt{khandker.islam@looop.co.jp} \\
  \And
  Nishat Anjum Bristy\textsuperscript{\affCmu,a} \\
  \texttt{nbristy@andrew.cmu.edu} \\
  \AND
  Faria Huq\textsuperscript{\affCmu,a} \\
  \texttt{fhuq@cs.cmu.edu} \\
  \And
  Priyeta Saha\textsuperscript{\affMiro,a} \\
  \texttt{priyeta.saha@miro.com} \\
  \And
  Syeda Nahida Akter\textsuperscript{\affCmu,\affNvidia,a} \\
  \texttt{sakter@nvidia.com} \\
  \And
  Arpita Saha\textsuperscript{\affPurdue,a} \\
  \texttt{saha119@purdue.edu} \\
  \And
  Tanzima Hashem\textsuperscript{\affBuet} \\
  \texttt{tanzimahashem@cse.buet.ac.bd} \\
}

\begin{document}

\maketitle

%
\begin{center}
\textsuperscript{\affUtor} University of Toronto, Toronto, ON, Canada \\
\textsuperscript{\affUwm} University of Wisconsin-Madison, Madison, WI, USA \\
\textsuperscript{\affVt} Virginia Tech, Blacksburg, VA, USA \\
\textsuperscript{\affLoop} Looop Inc., Tokyo, Japan \\
\textsuperscript{\affCmu} Carnegie Mellon University, Pittsburgh, PA, USA \\
\textsuperscript{\affMiro} Miro, Amsterdam, Netherlands  \\
\textsuperscript{\affNvidia} NVIDIA, Santa Clara, CA, USA \\
\textsuperscript{\affPurdue} Purdue University, West Lafayette, IN, USA \\
\textsuperscript{\affBuet} Bangladesh University of Engineering and Technology, Dhaka, Bangladesh \\
\vspace{0.8em}
\textsuperscript{a} \textit{Authors contributed equally to this research.} \\
\end{center}

\begin{abstract}
Advances in processing power, camera technologies, and mobile image analysis have made smartphones and other mobile devices, such as laptops, increasingly suitable for medical diagnosis and healthcare applications. Researchers have developed low-cost solutions for the early detection and monitoring of various health conditions, including eye and ENT diseases, malnutrition, heart rate variability, skin and oral conditions, and injuries, using images captured by non-medical devices such as smartphones and webcams. This survey examines existing research on mobile image-based medical diagnosis, with an emphasis on its potential to enable low-cost and accessible healthcare. We comparatively analyze state-of-the-art solutions across different healthcare application categories, examining their advantages and limitations. Based on this analysis, we identify desirable characteristics of mobile image-based diagnostic tools and highlight areas where existing approaches have made progress as well as areas requiring further research. We also discuss application-specific and common challenges and outline directions for future research. Overall, this study provides a comprehensive overview of mobile image-based healthcare solutions and their potential to support low-cost disease diagnosis and monitoring, particularly for underserved populations in remote and resource-constrained settings.

\end{abstract}


\section{Introduction}
Technological advancements in personal and mobile devices, particularly the availability of powerful integrated high-resolution cameras in smartphones and laptops, are reshaping healthcare delivery, making pervasive medical solutions more accessible. Despite progress in the healthcare sector, many individuals remain deprived of these benefits due to financial constraints, inadequate facilities, and a shortage of skilled workers. Mobile imaging addresses this gap by bringing essential diagnostic services within reach for people across diverse age groups and socio-economic backgrounds. While it cannot replace comprehensive clinical testing, it significantly improves access to initial screenings. In this survey, we provide a comprehensive review of existing research on mobile imaging solutions for medical diagnosis applications.



\begin{figure}[htb!]
    \centering
    \includegraphics[width=\linewidth]{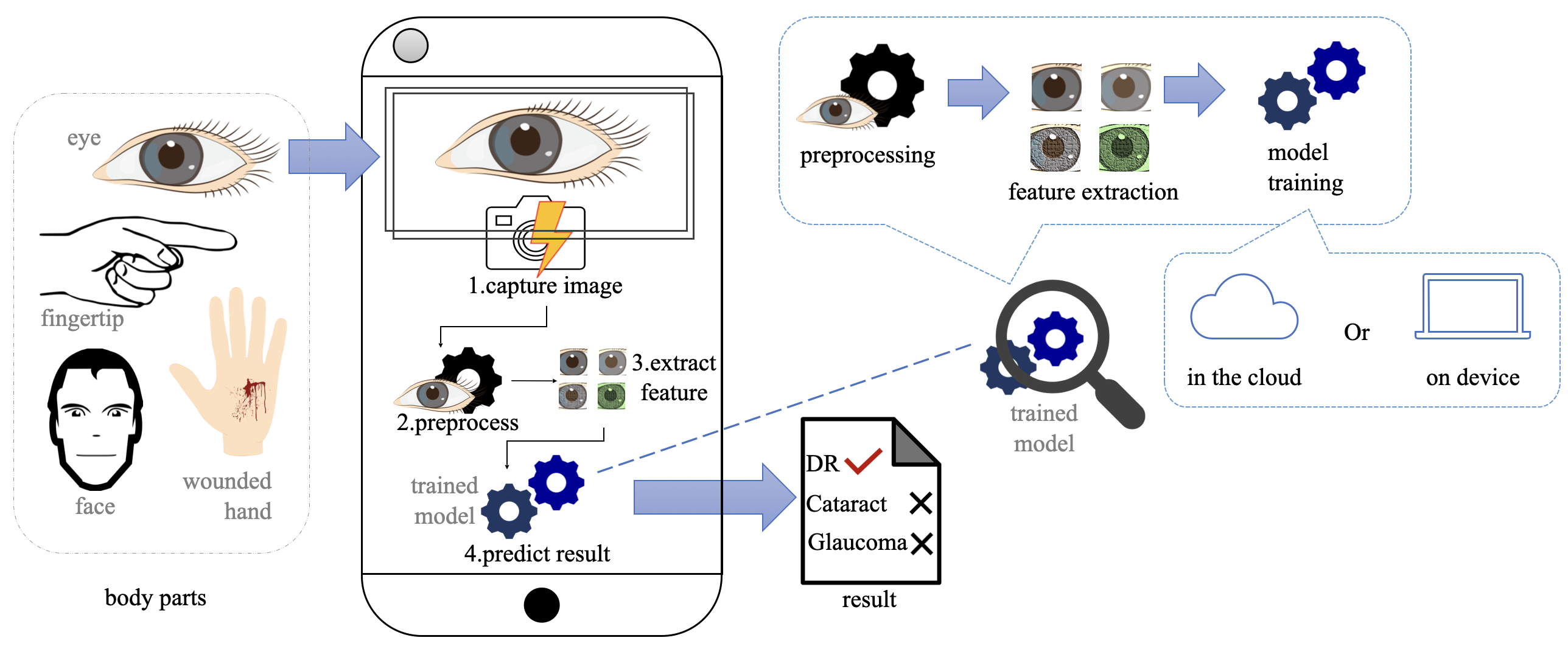}
    \caption{Mobile image analysis in healthcare applications}
    \label{fig:healthcare_applications}
\end{figure}

Images (including videos) captured with smartphones or laptop cameras can reveal crucial diagnostic cues from body parts such as the skin, eyes, or mouth, often unnoticed by the human eye. Unlike traditional imaging techniques like CT scans or MRIs, these devices provide an accessible means to detect a variety of health conditions, such as skin diseases, eye conditions, and early signs of malnutrition. By applying deep learning, machine learning, and other image analysis methods, meaningful features can be extracted from these images, enabling early detection of potentially life-threatening diseases. Our study highlights the techniques employed for early disease detection using image captures with non-medical devices, evaluates their strengths and limitations, and identifies key challenges that remain unaddressed.

Although several surveys on smartphone-based healthcare applications exist, their scope differs from ours. Most prior studies focus on specific diseases or applications without addressing the broader challenges of adopting mobile imaging solutions for medical diagnosis. For example,~\citep{kerst2020smartphone} reviews smartphone applications for managing depression, while~\citep{nejati2016smartphone} focuses on heart rate estimation, wound assessment and monitoring, and preliminary skin cancer detection. Similarly,~\citep{vashist2014commercial} explores tools for monitoring physiological parameters. In contrast,~\citep{mosa2012systematic} provides a broader perspective, with relatively less focus on disease diagnosis among its 15 categories of healthcare applications, which include drug references, clinical communication, Hospital Information Systems (HIS), and medical education. Recent advancements, however, have expanded the scope of smartphone applications by integrating technologies such as molecular analysis, biosensors, mathematical algorithms, microfabrication, 3D printing, and microfluidics. For example,~\citep{hernandez2019smartphone} demonstrates how these innovations enable smartphone applications in hematology, digital pathology, and rapid infectious disease diagnostics. Moreover,~\citep{litjens2017survey} examines the potential of smartphones in biomedical imaging, further highlighting their versatility in healthcare.

Our contributions in this survey paper are as follows:
\begin{compactitem}
    \item We provide a comprehensive review of medical diagnosis solutions using mobile images. 
    \item We present existing research in different categories and summarize the used techniques and pros-cons. Our comparative analysis facilitates the understanding of existing works in a common structure. 
    \item We identify both application-specific and common challenges of mobile imaging solutions for medical diagnosis and examine how these challenges have been addressed in the literature. 
    \item Finally, we provide direction for future researchers and draw attention to some important aspects that remain unexplored or less explored. 
\end{compactitem}

This paper is organized as follows. Section~\ref{sec:healthcareapp} categorizes existing works based on application types. Section~\ref{sec:analysis:app-specific} provides a comparative review of application-specific research. Section~\ref{sec:analysis:general} discusses common features and performs a general comparative analysis. Section~\ref{sec:futureworks} identifies challenges and future directions, and Section~\ref{sec:conclusion} concludes the survey.

\begin{figure}[htb!]
    \centering
    \includegraphics[width=\linewidth]{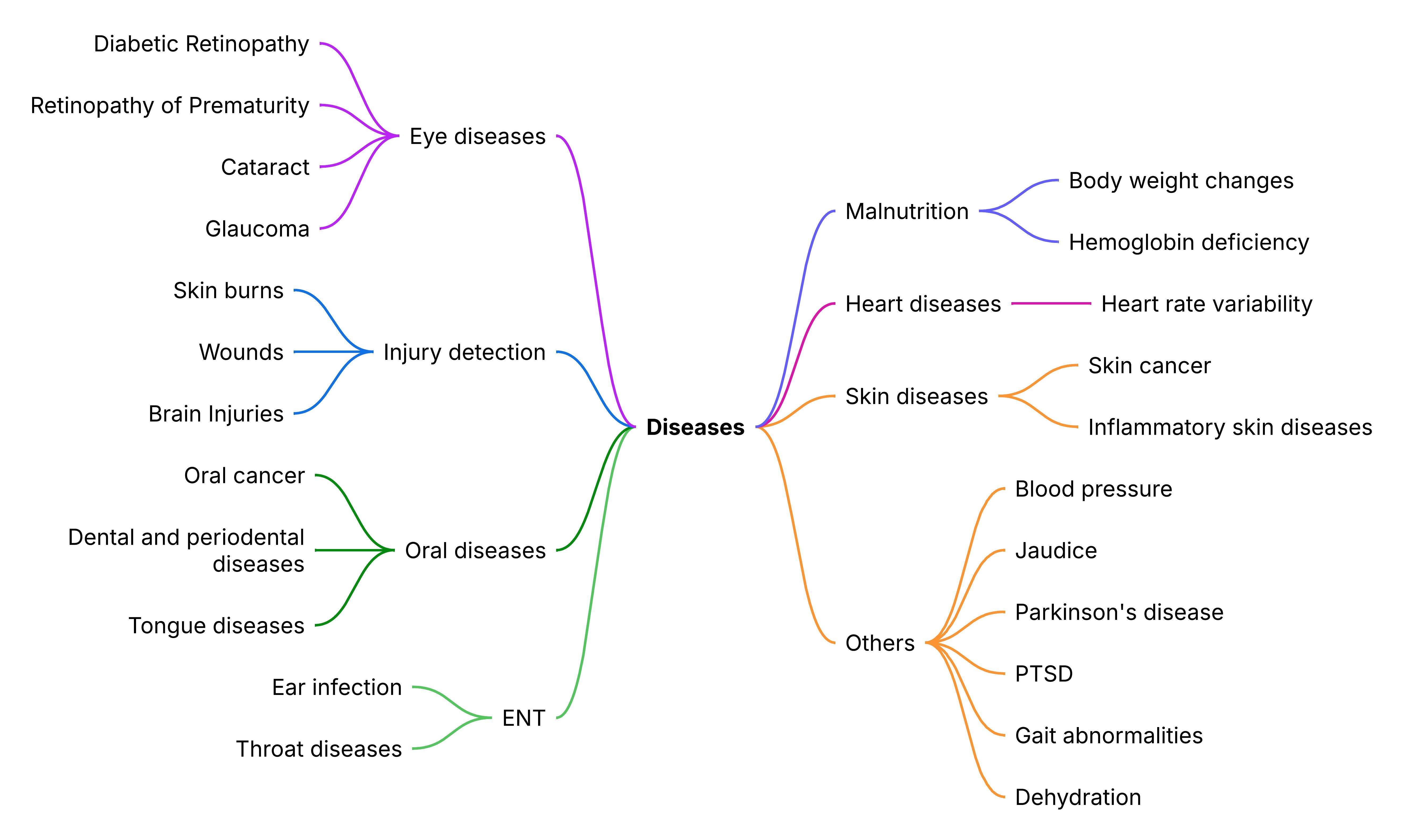}
    \caption{Taxonomy graph for included diseases categorized by application area}
    \label{fig:taxonomy_graph}
\end{figure}

\vspace{-1mm}
\section{Taxonomies}
\label{sec:healthcareapp}
\vspace{-1mm}

\label{subsec:archetypal_application_workflow}
In Fig.~\ref{fig:healthcare_applications}, we present a general mobile image analysis workflow for smartphone-based medical diagnosis applications. The process begins with capturing multimedia input of the relevant body part. Next, the input undergoes preprocessing, such as segmenting the region of interest (ROI), followed by feature extraction. These extracted features are then passed through a trained model to detect potential health issues. The model itself is trained using the extracted features from images of a specific body part. Since training a model can be computationally intensive, many researchers leverage cloud computing, where training data is uploaded to a central server equipped with powerful hardware capable of efficiently training large models. The trained model weights are then sent back to the device for use in healthcare applications aimed at early disease diagnosis. Alternatively, edge computing offers a decentralized approach, where both model training and disease inference are performed directly on the device, enabling greater autonomy and reducing dependency on external servers.


In Figure~\ref{fig:taxonomy_graph}, we categorize the diseases for which early diagnostic and monitoring tools have been developed, and present a taxonomy graph. In Table~\ref{tab:happ:summary}, for each category, we mention the information regarding what they are, their consequences, symptoms and standard diagnosis measures. We observe that most of the standard methods of diagnosis or measurement are expensive and require a person to visit a medical center. Most of the diagnoses are dependent on the expertise of a healthcare professional. We also observe that the diseases cause visible symptoms. Researchers have attempted to capture these symptoms using mobile image analysis to design low-cost, accessible diagnosis methods.

\changes{\emph{\textbf{Review Methodology.}} For selecting and narrowing down relevant publications for each categorized disease, we followed a systematic approach. Our representative search string in Google Scholar includes the following keywords:}

Smartphone based + \{disease name\} + (detection OR screening OR diagnosis OR evaluation) + \{disease specific keywords\}

\changes{For example, for the disease "Retinopathy of Prematurity (ROP)", our search string was "smartphone based (retinopathy of prematurity OR rop) (diagnosis OR screening) (fundus imaging)". The search results were then further assessed for eligibility on the grounds of i) in or out of scope for this review, ii) level of impact, iii) full text availability, and iv) selection of a single article from the same studies. Further expert judgement was used to focus on relevant publications that used state-of-the-art and novel approaches to solve difficult problems.}

\input{chap3.tex}

\section{Comparative Analysis - General} \label{sec:analysis:general}
In this section, we focus on the comparison of some common aspects, such as system architecture, models etc. of the previously discussed works. 

\subsection{Architecture} \label{Comparison-general-architecture}
\begin{figure}[htb!]
\centering
    \begin{subfigure}{.45\textwidth}
        \centering\includegraphics[width=\linewidth]{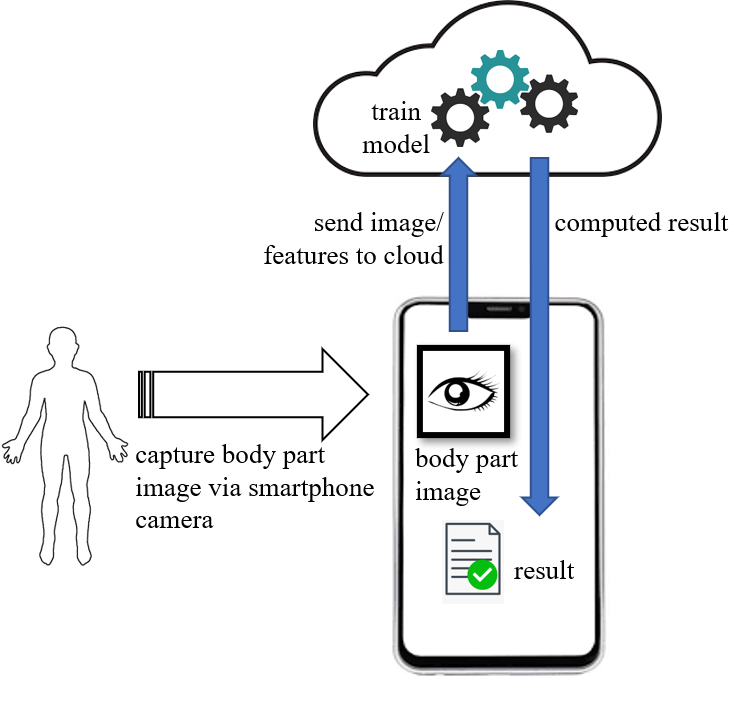}
        \caption{Cloud based architecture for smartphone-based image analysis}
    \end{subfigure}
\hfill
    \begin{subfigure}{.45\textwidth}
        \centering\includegraphics[width=\linewidth]{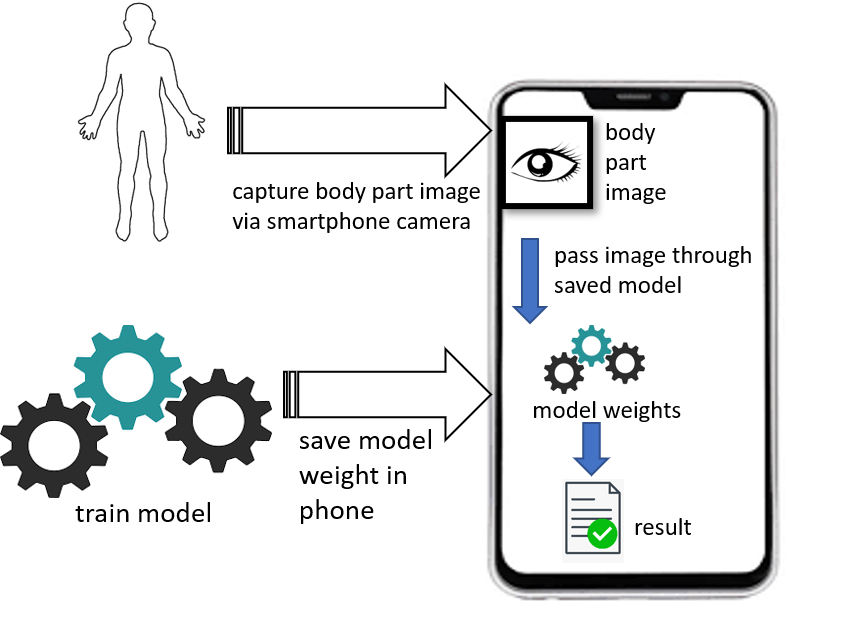}
        \caption{Offline architecture for smartphone-based image analysis}
    \end{subfigure}%
\caption{Architecture for image analysis for medical diagnosis}
\label{fig:arch}
\end{figure}
Few studies have thoroughly explored the system architectures for healthcare applications. Two primary types of architectures have been identified:
\begin{enumerate}
    \item \textbf{Cloud Based Architecture~\citep{rajalakshmi2018automated,kim2021comparison}:} In this architecture, models are stored in the cloud, requiring internet connectivity for classification tasks. Smartphones either send the raw input image or extract key features locally before transmitting them to the cloud. This approach allows models to be updated with new training data, improving their generalizability over time. However, reliance on connectivity and data privacy concerns are significant limitations.
    \item \textbf{Offline Architecture~\citep{natarajan2019diagnostic,sosale2020medios,martins2020offline}:} Here, model weights are stored locally on the smartphone, enabling all computations to occur on-device. This architecture is well-suited for low-connectivity environments and ensures data privacy. However, it requires lightweight models to accommodate the computational constraints of smartphones, potentially sacrificing some accuracy.
   
\end{enumerate}

\subsection{Models}

Most of studies that focused on detecting healthcare problems from digital images using image processing techniques, statistical and machine learning models were conducted in 2015. Over the past two decades, the popularity of deep learning models, especially CNNs (basic and advanced), has increased significantly. The usage of ANN (Artificial Neural Networks) began in 2002 but reached its peak in 2015.

We notice that signal analysis methods are particularly popular in heart rate variability (HRV) analysis due to their faster processing capabilities compared to machine learning and other sophisticated models. These models can also perform real-time HR detection using only a smartphone, with a minimal accuracy gap compared to ground truth values, further enhancing their popularity. 

SVM (Support Vector Machine) is widely used in domains such as skin cancer detection and inflammatory skin diseases. This is because SVMs can draw fine distinctions between similar skin diseases, with the model's hyperplane effectively separating the classifications, aiding in accurate predictions.

CNNs have been most widely used in diabetic retinopathy, skin cancer, and heart rate variability analysis. Initially, these models were used only for skin cancer and HRV detection, but with the advent of newer, more effective models, their applications have expanded across a broader range of healthcare domains.


\section{Challenges and Future Directions} \label{sec:futureworks}
\begin{itemize}
    \item \textbf{Architecture}: Current smartphone applications lack significant deployment of deep learning models, except in limited cases such as diabetic retinopathy, where offline and cloud-based architectures are utilized (see Section~\ref{Comparison-general-architecture}). The gap arises because smartphone processors, while improving, cannot match the computational power required for GPU-intensive deep learning tasks. Future exploration could involve designing personalized architectures in which models train locally on user-provided data and offer tailored feedback. Such architectures could revolutionize domains like blood pressure monitoring and diabetes management using smartphones.
    
    \item \textbf{Unified performance measure:} Healthcare applications within the same domain often lack standardized performance metrics. For example, in heart rate variability analysis, models use varying metrics such as discrepancy, RMSE, or error rates, which complicates cross-method comparison. Defining and adopting universal metrics for specific healthcare domains can improve consistency and reliability in evaluating models.
    
    \item \textbf{Explainability}: Deep learning models in healthcare primarily function as black boxes, offering little insight into their decision-making processes. Research to improve the interpretability of the model could lead to the design of more accurate input and increase user trust. Explainability tools such as SHAP~\citep{LundbergL17} or LIME~\citep{Ribeiro0G16} could help identify the features that drive model predictions. 
    
    \item \textbf{Environmental Robustness}: Image-based healthcare applications are often sensitive to environmental factors such as lighting or background noise, which can impact accuracy. Developing methods that eliminate environmental dependencies, such as advanced pre-processing or domain adaptation techniques, is a critical research direction.
    
    \item \textbf{User-centered application design}: Non-technical users may struggle to input clear and focused images into healthcare apps, leading to suboptimal results. Guided interfaces with features like on-screen alignment guides or stabilization prompts can enhance input quality and usability for non-expert users.
    
    \item \textbf{Benchmark dataset}: A lack of benchmark datasets across domains like skin disease detection and diabetic retinopathy hampers effective model comparison. Establishing standardized datasets would enable robust evaluation and foster competition, leading to better-performing models.
    
    \item \textbf{Computational constraints}: Deep learning models are often computationally expensive, making them unsuitable for low-end smartphones. Research into lightweight architectures using techniques like model pruning, quantization, or knowledge distillation can bridge this gap without sacrificing accuracy.

    \item \textbf{Ethical and Privacy Concerns}: With the increasing use of personal medical data for training and deploying models, ensuring data privacy and ethical use is critical. Research could focus on developing privacy-preserving techniques such as federated learning or differential privacy to ensure that sensitive user data is not exposed.

    \item \textbf{Real-World Validation}: One of the key challenges for mobile healthcare applications is testing them in real-world settings and validating their accuracy. Future work should focus on comparing the results of these diagnostic applications with findings from actual medical diagnoses to ensure their reliability and effectiveness in clinical practice. 
\end{itemize}

\section{Conclusion}\label{sec:conclusion}
With rapid advancements in mobile technology, smartphones have seamlessly integrated into our daily lives, enabling transformative innovations across multiple domains, including healthcare. Leveraging these capabilities, researchers have made remarkable progress in developing smartphone-based healthcare applications for early disease detection and monitoring. This survey provides a comprehensive review of the existing work in this domain, focusing on applications that utilize mobile image analysis.

We categorized healthcare applications, highlighting their desired characteristics and summarizing the advantages and limitations of current solutions. Our findings identify gaps in the field, such as the need for standardized performance metrics, robust architectures for real-world environments, and improved explainability. These insights aim to inspire further research to address these challenges effectively.

Moreover, this study serves as a guide for designing novel, image-based healthcare applications that can cater to emerging disease categories. By tackling the outlined challenges, the potential for creating accessible, low-cost diagnostic tools will significantly improve, particularly for underserved populations in remote and resource-constrained areas.

Although the field of smartphone-based image analysis in healthcare is quickly expanding with numerous studies each year, we found that several challenges remain unaddressed or inadequately addressed. Addressing these challenges will make low-cost smart healthcare more accessible to people in remote and underserved areas.

\bibliographystyle{unsrt}
\bibliography{references}

\appendix

    \section{Appendix}

    \begin{longtable}{|P{0.14\linewidth}|P{0.17\linewidth}|p{0.60\linewidth}|} 
        \caption{Summary of Various Healthcare Categories} \label{tab:happ:summary} \\
    \hline
    \textbf{Category} & \textbf{Disease} & \textbf{Description} \\
    \hline
    \endfirsthead

    \multicolumn{3}{c}%
    {\tablename\ \thetable\ -- \textit{Summary of Various Healthcare Categories [Continued]}} \\
    \hline
    \textbf{Category} & \textbf{Disease} & \textbf{Description} \\
    \hline
    \endhead

    \hline \multicolumn{3}{r}{\textit{Continued on next page}} \\
    \endfoot

    \hline
    \endlastfoot
\multirow{4}{=}{Eye Diseases} 
    & Diabetic Retinopathy (DR) & Diabetic Retinopathy is an eye disease that damages the retina due to diabetes mellitus, often leading to blindness in working adults. 

    \textbf{\textit{Symptoms:}} Gradually worsening vision, sudden vision loss, floaters, blurred or patchy vision, eye pain or redness, difficulty seeing in the dark.
    
    \textbf{\textit{Diagnosis:}} Diagnosed through stereoscopic fundus photographs.
    \\ \cline{2-3}
    & Retinopathy of Prematurity (ROP) & Retinopathy of Prematurity is an eye disease observed in premature infants and is a leading cause of blindness in children, preventable if treated early. 

    \textbf{\textit{Symptoms:}} Unusual eye movements, white pupils, vision loss.
    
    \textbf{\textit{Diagnosis:}} Diagnosed using fundus images.\\ 
    
    \cline{2-3}
    & Cataract  & Cataract is a common eye condition in the elderly that leads to poor vision and eventually blindness. 

    \textbf{\textit{Symptoms:}} Clouded or blurry vision, difficulty seeing at night, light sensitivity, seeing halos around lights, frequent changes in eyeglass prescription.
    
    \textbf{\textit{Diagnosis:}} Diagnosed through a visual acuity test.
    \\ \cline{2-3}
    & Glaucoma & Glaucoma is a cause of visual impairment and blindness that is often asymptomatic in its early stages, leading to undiagnosed cases until advanced. 

    \textbf{\textit{Symptoms:}} Intense eye pain, nausea, vomiting, red eyes, headaches, seeing rings around lights, blurred vision.
    
    \textbf{\textit{Diagnosis:}} Diagnosed using the Goldmann tonometer.    
    \\ \hline
\multirow{2}{=}{Malnutrition} 
    & Obesity & Obesity, characterized by excess body weight, is a major health condition leading to various associated health problems. 

    \textbf{\textit{Symptoms:}} Breathlessness, increased sweating, snoring, difficulty in physical activity, frequent tiredness, joint and back pain.
    
    \textbf{\textit{Diagnosis:}} Diagnosed through anthropometric measures such as weight, height, BMI, and waist-hip ratio.
    \\ \cline{2-3}
   & Hemoglobin Deficiency & Hemoglobin carries oxygen to the body’s tissues, and a deficiency may signal iron-deficient anemia caused by malnutrition, among other diseases.

    \textbf{\textit{Symptoms:}} Pale skin, brittle nails, pica syndrome (eating non-food items), lightheadedness when standing, shortness of breath, sore or inflamed tongue, mouth ulcers.

    \textbf{\textit{Diagnosis:}} Diagnosed through blood tests.
    \\ \hline

\multirow{2}{=}{Heart Diseases} 
     & Heart Rate Variability & Heart rate variability refers to the fluctuations in time intervals between adjacent heartbeats. An optimal heart rate variability is indicative of good health, adaptability, and resilience.

    \textbf{\textit{Symptoms:}} Fatigue, inflammation.
    
    \textbf{\textit{Diagnosis:}} Diagnosed via an electrocardiogram (ECG).    \\ \hline

\multirow{2}{=}{Skin Diseases} 
    & Skin Cancer & Skin cancer is a type of cancer that develops in the skin, the body's largest organ. It includes basal cell carcinoma, squamous cell carcinoma, and melanoma, with melanoma being the most unpredictable and dangerous.

    \textbf{\textit{Symptoms:}} A new spot on the skin, changes in the size, shape, or color of an existing spot, appearance of spots that are itchy or painful, non-healing sores that bleed or develop a crust.

    \textbf{\textit{Diagnosis:}} Diagnosed using a skin biopsy with histopathology.
    \\ \cline{2-3}
    & Inflammatory Skin Diseases & Inflammatory skin diseases, such as eczema, hives, psoriasis, atopic dermatitis, herpes, are prevalent in the modern population. These may be triggered by allergens, extreme temperatures and even stress. Despite their prevalence, these diseases are susceptible to misdiagnosis by inexperienced dermatologists. 

    \textbf{\textit{Symptoms:}} Smooth or scaly skin, skin may itch, burn or sting, flat or raised skin, skin redness, warmth in the affected skin area, blisters or pimples.

    \textbf{\textit{Diagnosis:}} Diagnosed using histopathology.
    \\ 
    \hline
\multirow{3}{=}{Injury Detection} 
    & Skin Burns & Skin burns are injuries caused by heat, electricity, radiation, chemicals, etc. Burns are classified into first, second, or third degree based on their severity.
    
    \textbf{\textit{Symptoms:}} Blisters, pain, swelling, red, white, or charred skin. 

    \textbf{\textit{Diagnosis:}} Diagnosed using a punch biopsy of burn tissue with subsequent histological analysis.
    \\ \cline{2-3}
    & Wounds & A wound can be considered an injury to living tissue caused by a cut, bump, or other impacts. It can be a symptom of other critical diseases such as diabetes or ulcers. 
    
    \textbf{}\textbf{\textit{Symptoms:}} Fever, swelling, redness, bleeding.
    
    \textbf{\textit{Diagnosis:}} Can be detected via naked eyes.
    \\ \cline{2-3}
    & Brain Injuries &Brain injuries are damage to the brain caused by external forces, such as a blow or impact, with concussions being a common result from rapid brain movement within the skull. 
    
    \textbf{\textit{Symptoms:}} Amnesia, headache, loss of consciousness, balance problems, nausea.
    
    \textbf{\textit{Diagnosis:}} Diagnosed using MRI or CT scans. 
    \\ \hline

\multirow{5}{=}{Oral Diseases} 
    & Oral Cancer & Oral cancer affects the mouth, including the lips, tongue, and other oral tissues, and is associated with high mortality rates, particularly if diagnosis is delayed.

    \textbf{\textit{Symptoms:}} Lump in neck, loose teeth, sore lips, difficult swallowing, unexplained weight loss.
    
    \textbf{\textit{Diagnosis:}} Diagnosed using tissue biopsy with histological assessment.
     \\\cline{2-3}
    & Tongue Diseases & Tongue diseases refer to conditions affecting the tongue, where changes in its coating and color can indicate underlying health issues, such as a purple tongue signaling high cholesterol or chronic bronchitis.
    
    \textbf{\textit{Symptoms:}} Pain, bumps on the surface of the tongue and changes to its color or texture.

    \textbf{\textit{Diagnosis:}} Diagnosed via histological examination.
    \\\cline{2-3}
    & Dental \& Periodontal Diseases & Dental and periodontal diseases refer to conditions affecting the teeth and the supporting structures, primarily caused by infections and inflammation of the gums and bone that surround and support the teeth
    
    \textbf{\textit{Symptoms:}} Severe pain, infection, tooth loss, inflammation in surrounding tissues. 

    \textbf{\textit{Diagnosis:}} Diagnosed via periodontal probing, measurement of clinical attachment loss, bone loss, cone beam computed tomography, quantitative polymerase chain reaction tests, etc.
         \\  \hline


    \multirow{2}{=}{ENT}  
    & Ear Infection & An ear infection is an inflammation of the ear, often caused by bacteria or viruses. Otitis media, a common type, can lead to hearing loss and impact a child's learning ability

    \textbf{\textit{Symptoms:}} Irritability, difficulty in sleeping or staying asleep, fever, fluid draining from ear(s), loss of balance, hearing difficulties, ear pain.
    
    \textbf{\textit{Diagnosis:}} Diagnosed using an otoscope.\\ \cline{2-3}
    & Throat Diseases & Throat diseases involve infections or inflammation of the throat, such as streptococcal pharyngitis (strep throat), a bacterial infection affecting the throat and tonsils.
    
    \textbf{\textit{Symptoms:}} Fever, pain when swallowing, sore throat, red and swollen tonsils, white patches or streaks of pus on the tonsils.

    \textbf{\textit{Diagnosis:}} Diagnosed using a throat culture test.
    \\  \hline

\multirow{5}{=}{Others}  
    & Blood Pressure (BP) & High blood pressure (hypertension) is a condition where the force of blood against the artery walls is consistently too high, leading to complications such as heart failure, vision loss, stroke, and kidney diseases. 

    \textbf{\textit{Symptoms:}} Headaches, blurred vision, chest pain.
    
    \textbf{\textit{Diagnosis:}} Measured using a cuff-based method (sphygmomanometer).
    \\ \cline{2-3}
    & Jaundice & Jaundice is a condition caused by an accumulation of bilirubin in the blood, often due to liver diseases, leading to yellowing of the skin, eyes, and mucous membranes.

    \textbf{\textit{Symptoms:}} Dark urine, abdominal pain, fever, nausea, yellow skin, pale stools, chills.

    \textbf{\textit{Diagnosis:}} Diagnosed using a bilirubin blood test.
    \\\cline{2-3}
    
    
    & Post-traumatic Stress Disorder (PTSD) & PTSD is a mental health disorder that occurs after experiencing or witnessing traumatic events, leading to flashbacks, nightmares, and severe anxiety.  

    \textbf{\textit{Symptoms:}} Flashbacks, nightmares and severe anxiety.

    \textbf{\textit{Diagnosis:}} Diagnosed using the CAPS-5, a 30-item structured interview.
    \\ \cline{2-3}
    &  Mental Health Diseases & Mental health conditions involve psychological disorders, including depression, anxiety, and other disorders affecting emotional well-being and mental functioning.
    
    \textbf{\textit{Symptoms:}} poor concentration, guilt or low self-worth, hopelessness, suicidal thoughts etc.
    
    \textbf{\textit{Diagnosis:}}  Diagnosed through therapy sessions and psychological assessments.
    \\ \cline{2-3}
    &  Gait Abnormalities &  Gait abnormalities refer to unusual walking patterns, often seen in conditions like cerebral palsy. 
    
    \textbf{\textit{Symptoms:}} Walking with head and neck bent, dragging feet, shuffling, or irregular jerky movements.
    
    \textbf{\textit{Diagnosis:}} Diagnosed using the Edinburgh Visual Gait Score (EVGS).
    \\ \cline{2-3}
    &  Dehydration & Dehydration occurs when the body loses more fluids than it consumes, leading to an imbalance in electrolytes and affecting normal bodily functions.
    
    \textbf{\textit{Symptoms:}} Headache, confusion, dizziness, fatigue, sunken eyes, dry mouth, dry cough.
    
    \textbf{\textit{Diagnosis:}} Diagnosed using plasma osmolality measurements.
    \\ \hline

    \end{longtable}


\end{document}

%% file: chap3.tex
\section{Comparative Analysis - Application Specific}\label{sec:analysis:app-specific}
In this section, we compare the existing research works for each category mentioned in Table~\ref{tab:happ:summary}. We thoroughly evaluate their image analysis pipeline: input, preprocessing steps, features, the classifiers employed, dataset, the output of the analysis, and performance. We identify the desired characteristics for a smart healthcare application and based on them, we check the pros and cons of the existing solutions. Some of these criteria are same for all categories; for example, if a solution does not require internet or skilled workers then it is suitable for people from destitute areas. On the other hand, some desired criteria are application specific; for example, if a diagnostic tool for eye diseases requires eye dilation then users will be not comfortable to use it. We summarize our findings in a comparison table for the health application sectors, where a considerable number of prominent studies are currently available. We observe that there is rarely a solution that fulfills all desired characteristics. Thus, our analysis and findings identify the limitations of the existing solutions and require further research.



\subsection{Eye Diseases}

\subsubsection{Diabetic Retinopathy}
In the last few years, there have been some significant works~\citep{rajalakshmi2018automated} on smartphone-based image analysis in diabetic retinopathy (DR) to make eye exams more affordable. These works have captured diagnosable retinal images~ or videos~\citep{mueller2020automated} using smartphone camera with the help of low-cost extensions, e.g. Remidio fundus on phone (FOP), Paxos Scope adapter, RetinaScope etc. These images were graded manually (ground truth) and in automated ways for DR screening. 
The automatic grading of mobile images has shown high (more than 80\%) specificity (SP) and sensitivity (SN) in most of these works. Among the automated techniques used, two are commercially available softwares for DR screening that use deep learning - (i) EyeArt\textsuperscript{TM} software~\citep{eyeart} and (ii) Medios AI~\citep{sosale2020medios}. \citep{rego2022implementation} assesses the diagnostic accuracy of image interpretations captured using EyeFundusScope. \citep{KASHYAP2020113} presents a cost-effective and lightweight mobile phone-based system for the early detection of diabetic retinopathy using artificial neural networks (ANN). Table~\ref{tab:DR:pros_cons} shows the pros and cons of existing works. 


\begin{table*}[h!]
\caption{Pros and cons of representative literature in \textit{diabetic retinopathy}}
\label{tab:DR:pros_cons}
\begin{tabular}{P{0.35\linewidth}P{0.1\linewidth}P{0.1\linewidth}P{0.1\linewidth}P{0.1\linewidth}P{0.1\linewidth}}
\toprule
    Specifications & \citep{rajalakshmi2018automated} & \citep{natarajan2019diagnostic} & \citep{sosale2020medios} & \citep{mueller2020automated} & \citep{kim2021comparison} \\ 
\toprule
    Requires internet connection for high performance computing & \cmark & \xmark & \xmark & N/S & \cmark\\ 
\midrule
    Requires skilled users & \cmark & \xmark & \cmark & \xmark & \xmark \\
\midrule
    Requires additional equipment & \cmark & \cmark & \cmark & \cmark & \cmark \\ 
\midrule
    Considers dark and noisy environment & \xmark & \xmark & \xmark & \xmark & \xmark \\    
\midrule
    Considers movement artifacts & \xmark & \xmark & \xmark & \xmark & \xmark \\    
\midrule
    Model integrated into a mobile application & N/A & \cmark & \xmark &  \cmark & N/A \\ 
\midrule 
    Performance validation w.r.t gold standard dilated eye test & \xmark & \xmark & \xmark & \xmark & \cmark \\ 
\midrule
    Requires pharmacological eye dilation & \cmark & \xmark & \cmark &  N/S & \cmark \\  
\bottomrule

\end{tabular}
\end{table*}

\subsubsection{Retinopathy of Prematurity}
Some works~\citep{lin2014smartphone,lekha2019mii,wintergerst2019non,goyal2019smartphone,patel2019smartphone} in the last decade have  validated that retinopathy of prematurity (ROP) in infants can be detected with reasonable accuracy using smartphone captured fundus images~\citep{wintergerst2019non} and videos~\citep{lin2014smartphone,lekha2019mii,goyal2019smartphone,patel2019smartphone}.~\citep{karakaya2020comparison} compared the image quality of different smartphone-based portable retinal imaging systems and their automatic DR detection accuracy using a deep learning framework.


\subsubsection{Cataract}
The use of eye images captured with smartphones has also been explored in automatic cataract detection~\citep{lee2015intelligent,lau2015mobile,kaur2015low,agarwal2019mobile,fuadah2015mobile,rana2017cataract,askarian2021detecting}. Lee et al. employed an optimized classifier based on feature elimination combined with a neural network for cataract classification~\citep{lee2015intelligent}. Similarly, Kaur et al. utilized a neural network for cataract detection~\citep{kaur2015low}. The studies in~\citep{fuadah2015mobile, agarwal2019mobile} applied K-nearest neighbors for this purpose. Lau et al. proposed using the red-eye effect as a detection method~\citep{lau2015mobile}; however, this approach remains underdeveloped. Only~\citep{askarian2021detecting} utilized a support vector machine for cataract detection after processing mobile images. 

\subsubsection{Glaucoma}
Glaucoma screening from mobile images has been studied in~\citep{boonarpha2018comparison,goh2018objective,kumar2019smartphone,li2020development,martins2020offline,lamonica2021remote, mrad2022fast} and concluded that mobile images have the potential to be used in glaucoma screening. Among them, only~\citep{martins2020offline} automated the screening through an interpretable pipeline that uses CNN. 


\subsection{Malnutrition}

\subsubsection{Body Weight}
Body Mass Index (BMI) is an widely used indicator of body weight. Some researchers have worked on BMI estimation from images. 
Jiang et al.~\citep{jiang2019body} collected 2950 frontal body photos, BMI measurements, and other pertinent data for a visual-body-to-BMI dataset. They extracted anthropometric features from images using a convolutional pose machine (CPM) and estimated BMI using support vector regression (SVR) and Gaussian processing regression (GPR). They show that body image to BMI prediction is feasible and usually more accurate than facial images. 
Jiang et al.~\citep{BMIMJiang2020} collected an RGB-D video dataset from 163 individuals using the Kinect sensor. They preprocessed the images using Skeleton joints mapping and estimated body volume from derived 3D data. To predict weight from images of dressed people, they developed two clothes models to exclude the effect of clothes on body volume computation. 
In contrast, Wen and Guo~\citep{wen2013face} targeted the correlation between facial features and BMI and used 14,500 facial images to conduct their experiments. They compute seven facial features using the active shape model (ASM) and ran SVR on those to produce BMI values. 
Affuso et al.~\citep{affuso2018method} measured obesity using body fat percentage. They took three images of 323  — front, back, and side profiles — and used the front curve and side curve to construct body form features. Then, they cluster participants using K-means and Calinski-Harabasz criterion. Finally, they trained their SVR model to predict body fat percentage. \citep{gadekallu2021identification} attempted to identify malnutrition and predict BMI using real-time facial image processing. \citep{HO2023112212} assessed the validity of 
exisiting image-assisted mobile nutrition apps.

\subsubsection{Hemoglobin Analysis}
Many alternatives to blood hemoglobin level testing for anemia diagnosis have been proposed in recent years~\citep{hasan2017,hasanReview2021}.
HemaApp, a smartphone app by Wang et al.~\citep{wang2016hemaapp}, measures blood hemoglobin concentration noninvasively. They attached IR light sources of different wavelengths in hardware. The app records fingerlight reflection videos for each light source. It turns these videos into RGB time-series waveforms, identifies features, and trains SVM regression models based on blood test hemoglobin levels for each wavelength. HemaApp is effective in anemia classification (with 85.7\% sensitivity and 76.5\% specificity). This work was extended to HemaApp V.2, by using the smartphone's back camera and white flash LED~\citep{wang2017noninvasive}, and achieved an RMSE of 1.27.
Mannino et al. went one step further and developed a standalone app to measure blood hemoglobin levels without extra equipment or smartphone attachment~\citep{mannino2018smartphone}. They used this app to prompt their subjects to take photos of their fingernails and select the ROI. Fingernail and skin color data were extracted from these ROIs and fed into a robust multi-linear regression algorithm with bi-square weighting. This system diagnosed anemia with 92\% sensitivity and 76\% specificity.  
Hasan et al.~\citep{hasan2018smarthelp} recorded 10-second fingertip videos, generated color maps of pixel color (RGB) intensity for each frame, and then segmented the images into blocks to capture pixel value changes. They trained their ANN model with block mean pixel values. The model predicted hemoglobin levels with 94\% sensitivity and 96\% specificity on their test dataset.  
Hasan et al.~\citep{hasan2018smartphone} proposed another approach that converted RGB pixel intensities into HSV, Lab, and gray color spaces. They predicted hemoglobin levels from color space values using Partial Least Squares (PLS). \citep{das2023smartphone} proposed a system which can efficiently access blood haemoglobin levels non-invasively by observing color changes in nail-bed.~\cite{chen2024real} introduced a novel smartphone-based system for non-invasive hemoglobin measurement via eyelid imaging, utilizing deep learning algorithms such as EGE-Unet and DHA (C3AE) models. This system demonstrated greater accuracy and efficiency compared to traditional methods and manual assessments.

\subsection{Heart Rate Variability}
There are two standard methods for measuring heart rates using smartphone applications: (i) from fingertips \citep{sukaphat2016heart,zaman2017novel}, which has evolved from a contact-based method to a non-contact based method \citep{khong2021contact} (ii) from facial images and videos, which is essentially a non-contact method. Following, we analyze existing studies on these two methods. Table~\ref{tab:heart:pros_cons} discuss the pros and cons of those works.

\subsubsection{Fingertip-based Heart Rate Measurement}
Sukaphat et al. measured heart rate using the image of a single fingertip \citep{sukaphat2016heart}. This study employed the mean value of RGB color channel histograms to determine heart rate frequency. This approach matched a digital pressure monitor by 0.57±\% when tested on 10 people.  
Conversely, Zaman et al. used successive fingertip images to calculate heart rate \citep{zaman2017novel} using two methods. In the first method, the authors applied edge detection and smoothing, and then, measured the change of the fingertip curve lines with time, which can be converted into time and frequency domain graphs to detect heart rate. In the second method, the authors used the changes in the image intensity of the red channel of the successive fingertip images to compute heart rate. 

\begin{table*}[htbp]
\caption{Pros and cons of representative literature in \textit{heart rate variability}}
\label{tab:heart:pros_cons}
\begin{tabular}{P{0.15\linewidth}P{0.05\linewidth}P{0.05\linewidth}P{0.05\linewidth}P{0.05\linewidth}P{0.05\linewidth}P{0.05\linewidth}P{0.05\linewidth}P{0.05\linewidth}P{0.1\linewidth}P{0.05\linewidth}}

\toprule
    Specifications & \citep{sukaphat2016heart} & \citep{zaman2017novel} & \citep{poh2010non} & \citep{pursche2012video} & \citep{kwon2012validation} & \citep{monkaresi2014machine} & \citep{yu2015dynamic} & \citep{osman2015supervised} & \citep{balakrishnan2013detecting} & \citep{shan2013video} \\
\toprule
    Requires internet connection for high performance computing & \xmark & \cmark & N/S & N/S & \xmark & N/S & N/S & N/S & N/S & N/S\\ 
\midrule
    Requires skilled users & \xmark & \xmark & \xmark & \xmark & \xmark & \xmark & \xmark & \xmark & \xmark & \xmark\\
\midrule
    Requires additional equipment & \xmark & \cmark & \xmark & \xmark & \xmark & \xmark & \xmark & \xmark & \xmark & \xmark\\    
\midrule
    Considers dark and noisy environment & \xmark & \xmark & \xmark & \xmark & \xmark & \cmark & \xmark & \cmark & \cmark & \cmark\\    
\midrule
    Considers movement artifacts & N/A & N/A & \cmark & \xmark & \xmark & \cmark & \cmark & \cmark & \xmark & \xmark\\    
\midrule
    Model integrated into a mobile application & \cmark & \cmark & \xmark & \xmark & \cmark & \xmark & \xmark & \xmark & \xmark & \xmark\\ 
\midrule
    Support mobile devices/ cameras with variant configuration & \xmark & \xmark & \xmark & N/S & \cmark & \xmark & \xmark & \xmark & \xmark & \xmark\\    
\midrule
    HR estimation for multiple person at a time & \xmark & \xmark & \cmark & \xmark & \xmark & N/S & N/S & \xmark & N/S & N/S\\    
\midrule
    Performance validation w.r.t ground truth & \cmark & \cmark & \cmark & \cmark & \cmark & \cmark & \cmark & \cmark & \cmark & \cmark\\
\midrule
    Requires skin to be visible & \cmark & \cmark & \cmark & \cmark & \cmark & \cmark & \cmark & \cmark & \xmark & \xmark\\
\midrule
    Requires RGB color images/ videos & \cmark & \cmark & \cmark & \cmark & \cmark & \cmark & \cmark & \cmark & \xmark & \xmark\\
\bottomrule

\end{tabular}
\end{table*}

\subsubsection{Facial Images based Heart Rate Measurement}
There are currently two remote methods for measuring heart rate (HR) from face videos: color intensity-based and motion-based. Photoplethysmography (PPG) signals captured by a camera are utilized in light-intensity based methods that measure variations in blood volume by detecting changes in light reflection or transmission during the cardiovascular pulse cycle. This idea was first introduced in 2008 by Verkruysse et al.~\citep{verkruysse2008remote} and later got adopted by~\citep{poh2010non, pursche2012video, kwon2012validation, yu2015dynamic, monkaresi2014machine}. 
The heartbeat signal is extracted using blind source separation (BSS) on facial color changes. The ROIs are selected from the videos, although the relative importance of the ROIs is still debatable. Independent Component Analysis (ICA) is the most common BSS method used to derive the face BVP signal.
Initially, Poh et al. \citep{poh2010non} and Kwon et al. \citep{kwon2012validation} used the second component of the ICA (i.e., the green channel) and attained RMS error less than 5 bpm compared to Flexcomp finger BVP sensor and average error rates of 1.04\% compared to ECG, respectively. Later, other studies~\citep{pursche2012video, yu2015dynamic, monkaresi2014machine} have analyzed all ICA components and improved performance. Though ICA and Principal Component Analysis (PCA) work experimentally well, they consider an observed signal to be a linear combination of multiple sources, which conflicts with the Beer-Lambert law that states that the intensity of light reflected through facial tissue is a nonlinear function of distance \citep{xu2014robust, wei2012automatic}. Some other dimensionality reduction algorithms used to extract face BVP signals are linear discriminant analysis, locally linear embedding, and manifold learning methods such as Isomap, Laplacian Eigenmap (LE) \citep{zhang2004principal}. Wei et al. \citep{wei2012automatic} investigated nine such algorithms on RGB color channels and found that Laplacian Eigenmap extracted BVP information best.

In contrast, HR detection methods based on motion originated from ballistocardiogram. Ballistocardiography is a technique used to record the repetitive motions of the human body in the form of ballistocardiogram 
signals. Balakrishnan et al. \citep{balakrishnan2013detecting} were the were the first to propose analyzing head motions during remote HR detection from face videos. They used PCA to project the pulse generated from head motion to 1D signals. This method yielded an error rate of 2.2\% when compared to ECG outputs, indicating that methods analyzing facial video provide superior performance.  Shan et al. ~\citep{shan2013video} chose a single feature point from the head (instead of considering all points in the head ~\citep{balakrishnan2013detecting}) and demonstrated that ICA is more consistent and accurate than PCA..

\subsection{Skin Diseases}
\subsubsection{Skin cancer}

Hameed et al. compiled studies up to 2016 on skin cancer using image analysis techniques \citep{hameed2016comprehensive}. In those studies, images were preprocessed by scaling, filtering, cropping, and hair removal. Color-based, region-based, and threshold-based segmentation approaches were employed. The most popular classifier was SVM, followed by ANN and then KNN. 
Later, Brinker et al. summarized CNN-based studies for diagnosing skin cancers through 2018 \citep{brinker2018skin}. Some of those studies employed CNN as a feature extractor, but the majority used transfer learning with CNN for end-to-end learning. 
Esteva et al. trained a deep CNN to classify images with similar accuracy to a dermatologist \citep{esteva2017dermatologist}. Later, Brinker et. al employed a pre-trained ResNet50 model that outperform dermatologists \citep{brinker2019deep}. CNN was also used to diagnose and categorize skin cancer in several recent research \citep{hajgude2019skin,nahata2020deep,zhang2020skin,ali2021enhanced}. 
Additionally, Kumar et al. used fuzzy C-means clustering for segmentation and trained ANN using a meta-heuristic algorithm, differential evolution, to successfully detect skin cancer \citep{kumar2020ann}. All these studies were desktop-based, however, SkinScan{\copyright}, a light-weight portable library for the automatic detection of melanoma suitable for smartphones and handheld devices, was developed in 2011,  \citep{wadhawan2011skinscan} and Kalwa et al. recently developed an end-to-end smartphone application for identifying melanoma with 88\% accuracy \citep{kalwa2019skin}. 
Table~\ref{tab:skincancer:pros_cons} highlights the pros and cons of the existing research works. Note that, most of these works have not evaluated the performance of their skin cancer identifying models for smartphone captured images.  

\begin{table*}[htbp]
\caption{Pros and cons of representative literature in \textit{skin cancer}}
\label{tab:skincancer:pros_cons}
\begin{tabular}{P{0.35\linewidth}P{0.1\linewidth}P{0.1\linewidth}P{0.1\linewidth}P{0.1\linewidth}P{0.1\linewidth}}

\toprule
    Specifications & \citep{abuzaghleh2014automated} & \citep{hajgude2019skin} & \citep{kalwa2019skin} & \citep{brinker2019deep} & \citep{kumar2020ann} \\ 
\toprule
\midrule
    Requires skilled users & \xmark & \xmark & \xmark & \xmark \\
\midrule
    Requires additional equipment  & \xmark & \xmark & \cmark & \xmark & \xmark\\    
\midrule
    Considers dark and noisy environment & N/S & N/S & N/S & \xmark & \xmark\\    
\midrule
    Model integrated into a mobile application & \cmark & \xmark & \cmark & \cmark & \xmark \\ 
\midrule
    Supports devices with variant configurations & N/S & N/S & \xmark & \xmark & \xmark\\    
\midrule
    Performance validation w.r.t ground truth & \cmark & \xmark & \cmark & \cmark & \cmark\\    
\bottomrule
\end{tabular}
\end{table*}

\subsubsection{Inflammatory Skin Diseases}
Several studies~\citep{gupta2019analysis,alamdari2016detection,yasir2015skin} examined the diagnosis procedure using a mobile application. 
Gupta et al. utilized Gaussian Mixture Models due to their greater computational efficiency and resource-friendliness, making them suitable for smartphones with resource constraints, unlike heavier deep learning models~\citep{gupta2019analysis}. 
Alamdari et al. studied low-resolution input images fed to the machine learning models as images from smartphones are expected to be low-resolution \citep{alamdari2016detection}. 
Yasir et al. surveyed the doctors in a developing country where skin diseases are widely prevalent and developed a system to detect the nine most common diseases by  studying the benefits and usability of such an application \citep{yasir2015skin}. 
Some researchers developed computer-based systems that use digital skin images from different websites as input, which can be extended easily to mobile applications using cross-platform developments \citep{alenezi2019method, hasija2017automated, abdul2012dermatology}. Many of these works performed heavy preprocessing using various image processing techniques to reduce noise and sensitivity to lighting conditions and finally to segment the affected areas from the image before feeding them into the model. Mostly two types of features, color and texture, were extracted before the classification task indicating that these features aid the model in giving accurate predictions. Although there was no significant usage of statistical models as classifiers, lightweight machine learning models were mostly used with SVM gaining popularity among them. 
In a few studies such as \citep{arifin2012dermatological} and \citep{yasir2015skin}, external features such as elevation were input to the system by the user along with the skin image, but the majority of the works in this domain used only skin image as input. 
In some studies \citep{alam2016automatic,alamdari2016detection}, the system uses metrics like EASI and other scoring algorithms to determine the severity of the illness, which might assist users to comprehend the gravity of their condition and decide what steps to take next. Table~\ref{tab:inflamskin:pros_cons} summarizes the pros and cons of existing works related to inflammatory skin diseases. 

\begin{table*}[htbp]
\caption{Pros and cons of representative literature in \textit{inflammatory skin diseases}}
\label{tab:inflamskin:pros_cons}
\begin{tabular}{p{0.45\linewidth}p{0.05\linewidth}p{0.05\linewidth}p{0.05\linewidth}p{0.05\linewidth}p{0.05\linewidth}p{0.05\linewidth}p{0.05\linewidth}}

\toprule
    Specifications & \citep{arifin2012dermatological} & \citep{gupta2019analysis} & \citep{alamdari2016detection} & \citep{alenezi2019method} & \citep{alam2016automatic} & \citep{de2015design2} & \citep{wu2020deep} \\ 
\toprule
    Requires internet connection for high performance computing & \xmark & \xmark & \xmark & \xmark & \xmark & \xmark & N/S\\ 
\midrule
    Requires skilled users to operate the system & \cmark & \xmark & \xmark & \xmark & \xmark  & \xmark& \xmark\\
\midrule
    Requires additional equipment & \xmark & \xmark & \xmark & \xmark & \xmark &\xmark & \xmark\\ 
\midrule
    Requires additional information input by a trained professional  & \xmark & \xmark & \xmark & \xmark  & \xmark & \xmark & \cmark\\
\midrule
    Considers dark and noisy environment & \xmark & \xmark  & \xmark & \xmark  & \xmark & \cmark & \xmark\\    
\midrule
    Accounts for other objects present in the image except skin & \cmark & \xmark  & \cmark & \xmark & & \cmark & \xmark\\
\midrule
    Able to identify multiple diseases or occurrence of a disease from a single image 
    & \cmark & N/S & \cmark & N/S & \cmark & \xmark & N/S\\  
\midrule
    Model integrated into a mobile application or a computer system & \xmark & \xmark & \xmark & \cmark & \xmark & \xmark & \cmark\\ 
\midrule
    Device specific performance validation & N/S & \xmark  & \xmark  & \cmark & N/S & \xmark & \xmark\\    
\bottomrule
\end{tabular}
\end{table*}

\subsubsection{Other}
Chan et. al made a comprehensive compilation of machine learning applications in dermatology \citep{chan2020machine}. In most cases, the models used for classifying dermatological diseases (e.g. melanoma, non-melanoma skin cancer) were as accurate (based on metrics like sensitivity, specificity, AUC) as the diagnoses done by board-certified dermatologists. However, due to the lack of external validation, these models often might not yield expected results for other real-life datasets. The review also highlighted the fact that existing mobile applications for classifying skin diseases are still not accurate enough to be clinically approved although they can be highly useful in solidifying the diagnosis process for dermatologists. In addition, some crucial limitations of machine learning (e.g. black-box nature of ML algorithms, dependency on input quality, algorithmic bias, lack of inclusivity, clinical regulations) were discussed in \citep{chan2020machine}.

\subsection{Injury Detection}
\subsubsection{Skin Burns}
Pande et al. proposed a method for categorizing burn levels and recommending first-aid treatments based on these levels~\citep{pande2014automated}. Chauhan et al. utilized ResNet50 to classify burnt body parts, aiming to automate the early assessment of the total body area affected by burns~\citep{chauhan2018using}. Abubakar et al. conducted pioneering work on classifying skin-burn images across different racial groups, specifically focusing on Caucasian and African skin types~\citep{abubakar2020assessment}. More recently, an increasing number of studies have employed CNNs and transfer learning techniques (e.g., pretrained ResNet, VGG16) to distinguish between burnt and injured skin~\citep{abubakar2020comparison} or to assess burn depth~\citep{tran2016degree,abubakar2020burns}. Table~\ref{tab:skinburn:pros_cons} summarizes the advantages and limitations of the various studies on skin burn detection and classification.


\begin{table*}[htbp]
\caption{Pros and cons of representative literature in \textit{skin burn}}
\label{tab:skinburn:pros_cons}
\begin{tabular}{P{0.35\linewidth}P{0.1\linewidth}P{0.1\linewidth}P{0.1\linewidth}P{0.1\linewidth}P{0.1\linewidth}}

\toprule
    Specifications & \citep{pande2014automated} & \citep{jiao2019burn} & \citep{abubakar2020assessment} & \citep{abubakar2020burns} & \citep{liu2021framework}  \\
\toprule
    Requires internet connection for high performance computing & \cmark & \cmark & \cmark & \cmark & \cmark \\  
\midrule
    Requires skilled users & \xmark & \xmark & \xmark & \xmark & \xmark \\
\midrule
    Requires additional equipment & \xmark & \xmark & \xmark & \xmark & \xmark \\   
\midrule
    Considers dark and noisy environment & N/S & N/S & N/S & N/S & N/S \\ 
\midrule
    Model integrated into a mobile application & \xmark & \xmark & \xmark & \xmark & \xmark  \\ 
\midrule
    Supports devices with variant configurations & \xmark & \xmark & \xmark & \xmark & \xmark \\  
\midrule
    Performance Validation w.r.t ground truth & \cmark & \cmark & \cmark & \cmark & \cmark \\     
\bottomrule
\end{tabular}
\end{table*}

\subsubsection{Wound Detection}
A variety of wound assessment systems have been developed to aid diabetic patients and monitor healing. In~\citep{bhelonde2015flexible, wang2015smartphone}, an accelerated mean-shift algorithm was used for wound segmentation, followed by a connected region detection method to determine wound boundaries. The healing state was then measured using red-yellow-black color rating models based on trend analysis of patient time records. Another system for diabetic wound assessment employed the K-means shift algorithm~\citep{dalya2016design}.

More recently, Shenoy et al. developed Deepwound, a multi-label CNN ensemble capable of classifying wound images using only pixel data and labels~\citep{shenoy2018deepwound}. Kaile et al. introduced a smartphone-based near-infrared imaging tool to quantify tissue oxygenation, a critical factor in wound healing. Changes in oxygen levels were shown to correlate with healing status~\citep{kaile2019low}.

Furthermore, Lau et al. presented a real-time pressure injury wound detection and staging classification system using YOLOv4~\citep{lau2022artificial}. This application highlights the growing potential of real-time machine learning models in clinical wound assessment.

\subsubsection{Brain Injury Detection}
Mariakakis et al. developed a smartphone-based system for detecting concussions by assessing pupillary light reflex~\citep{Mariakakis2017PupilScreenUS}. Using a smartphone flashlight to stimulate the eyes and recording the response via video, they employed a fully convolutional network (FCN) to track pupil diameter changes over time. This analysis allowed extraction of clinically relevant metrics, including constriction amplitude, percentage, and velocity. To minimize bias, the dataset included various iris colors, and conservative cropping excluded skin areas. The app, tested on individuals with mild traumatic brain injury (mTBI), offered a low-cost and portable pupillometer.


Additionally,~\cite{ko2016smartphone} introduced a smartphone-enabled optofluidic platform for the rapid detection of brain-derived exosomes, offering a faster and non-invasive method to identify potential biomarkers for concussion recovery compared to traditional techniques.

\subsection{Oral Diseases}
\subsubsection{Oral Cancer}
Several techniques have been employed in the literature to detect oral cancer from intraoral and whole-cavity imaging. Previously, Anuradha et al. proposed a system to detect benign or malignant tumors using the SVM classifier \citep{anuradha2013comparison}. Recently, smartphone-based probes have shown remarkable results in detecting oral cancer using CNN \citep{firmalino2018first, uthoff2018development, uthoff2018point}. These studies demonstrated sensitivity and specificity ranging from 81.25\% to 94.94\%, compared to specialists. 
"\citep{lin2021automatic} proposed a new image-capturing method for consistent lesion positioning and used HRNet for oral cancer classification. \citep{SHAH202270} developed a novel automatic oral cancer detection algorithm to identify and differentiate premalignant lesions from buccal cavity images, aiding in the early detection of oral cancer. \citep{fu2020deep} used a cascaded CNN to detect oral cavity squamous cell carcinoma (OCSCC) from photographic images.

\subsubsection{Tongue Diagnosis} 
The review paper by Jung et al.~\citep{jung2012review} summarizes studies up to 2012, where most research relied on data collected using external hardware for tongue diagnosis. Since then, the required devices have evolved from relying on external equipment to using only smartphones.

Zhang et al.~\citep{zhang2013tongue} applied SVM and KNN algorithms to confirm the link between human health and tongue color. They categorized tongue images into healthy and diseased with 91.99\% accuracy, and detected 11 diseases with 70\% accuracy from a dataset of 1,045 tongue images (143 healthy and 902 diseased). In this study, tongue images were assigned to one of 12 color gamuts for classification. Another study~\citep{zhang2013new} introduced 
or capturing tongue images, utilizing an optimized Canny algorithm to detect tongue outlines, followed by statistical analysis for automated diagnosis. Ryu et al.~\citep{ryu2014tonguedx} developed a tongue color calibration method to correct smartphone-captured tongue images by adjusting the white balance of teeth color, compensating for the effects of surrounding light exposure.

In contrast, Hu et al.~\citep{hu2014automatic, hu2016color, hu2019automated} proposed an SVM-based approach for estimating lighting conditions, predicting a color correction matrix based on the color difference between images taken with and without a flashlight. Kanawong et al.~\citep{kanawong2017tongue} conducted computer-aided tongue image analysis based on traditional chinese medicine.

Most recently, Huang et al.~\citep{huang2023tonguemobile} developed an automated tongue diagnosis system using Mask R-CNN for tongue-image segmentation and ResNeXt for tongue-coating color classification. Their mobile app implementation achieved significant improvements in accuracy for non-invasive disease assessment in traditional Chinese medicine.
 
\subsubsection{Dental Diseases}
A few notable works, such as~\citep{choi2018research, zhang2020smartphone}, have focused on caries detection using smartphone images. Zhang et al.~\citep{zhang2020smartphone} proposed an automated smartphone application for detecting early childhood caries, which was evaluated by trained dentists but still requires improvements to match their performance levels.  They employed a Single Shot Multibox Detector with a MobileNet-v2 backbone as the classifier, which is lightweight and well-suited for resource-constrained smartphones, enabling faster, real-time inference without requiring internet connectivity. On the other hand, Choi et al.~\citep{choi2018research} addressed lighting effects in images, enhanced image quality, and applied adaptive hue thresholding to extract the teeth region. They also used erosion and dilation techniques to improve image accuracy. Their Android application took user input (images) and performed all computations on a cloud server, requiring an internet connection.


In 2019, Askarian et al.~\citep{askarian2019smartphone} studied the use of smartphones for detecting periodontal diseases. They collected 30 images (15 diseased and 15 healthy) using an iPhone X rear camera with an add-on gadget to reduce refraction effects and ambient light interference. The images were preprocessed using color tuning and correction to reduce noise and extract the gum region (ROI). A gum color gamut was designed to represent all possible gum surface colors, and features were extracted from the CEILAB color space. SVM was used as the classifier, achieving an average accuracy of 94.3\%, specificity of 93\%, and sensitivity of 92.6\%.

Finally, You et al.~\citep{you2020deep} introduced a deep learning-based AI model for detecting dental plaque on primary teeth. Users uploaded smartphone-captured images, and the model demonstrated clinically acceptable diagnostic accuracy, achieving a mean intersection-over-union (MIoU) of 0.736, comparable to the performance of an experienced pediatric dentist.

\subsubsection{Miscellaneous Oral Diseases}
To enable users to regularly inspect their oral health independently, Liang et al. developed an interactive mobile app named OralCam~\citep{liang2020oralcam}. They employed a Multi-Task Learning approach with a single deep CNN model to extract features from input images capturing five different oral conditions: periodontal diseases, caries, soft deposit, dental calculus, and dental discoloration. In the second stage of their algorithm, another deep CNN model was used to locate bounding boxes for each finding. To train their model, the researchers prepared an in-house dataset of 3,182 oral cavity images annotated by dental experts. The dataset also included information about users’ living habits, history of pain, and bleeding. In addition to traditional evaluation metrics like accuracy and sensitivity, they incorporated feedback from end users and board-certified dentists to improve the app’s robustness. This work demonstrates significant promise in creating accessible and functional mobile health solutions.

\subsection{ENT}

\subsubsection{Ear Infection Detection}
Otitis Media (middle-ear infection) can be detected using a smartphone with a plug-in otoscope device. Huang et al.~\citep{otoscope_1} captured images using an otoscope and applied an active contour algorithm to segment the region of interest (ROI). Different Otitis Media cases were then classified using a Depth-First Search algorithm. The images were transmitted to a server for analysis. Myburgha et al.~\citep{otoscope_2} used a neural network to classify normal and Otitis Media cases. They employed a portable digital otoscope to capture images, which were saved and displayed on a smartphone before being processed on a dedicated server. In a more recent development, Chen et al.~\citep{chen2022smartphone} designed a smartphone-adapted deep learning model for detecting and diagnosing middle-ear diseases. They leveraged transfer learning and lightweight CNNs such as MobileNet, making the system highly efficient for use on mobile devices.

\subsubsection{Throat Diseases}
The use of smartphones for detecting Streptococcal Pharyngitis (strep throat) has been explored in~\citep{askarian2019novel, yoo2020toward}. Askarian et al. compiled high-resolution throat images (4032×3024 pixels) with metadata like age and sex, while Yoo et al. used low-resolution images (256×256 pixels) without metadata. Yoo et al. addressed the issue of limited datasets by collecting 131 pharyngitis and 208 normal images, classified by clinicians, and augmented the data using CycleGAN to generate paired normal and pharyngitis cases. Both studies used balanced training datasets: 20 strep/20 healthy images in~\citep{askarian2019novel} and 1500 strep/1500 healthy images in~\citep{yoo2020toward}.

For preprocessing, Askarian et al. applied color correction and intensity-based segmentation to extract throat ROIs, leveraging features like redness and swollen tonsils. Yoo et al. skipped explicit preprocessing or feature extraction. Classifier choice also differed: Askarian et al. employed KNN for its lightweight nature, achieving 93.75\% accuracy (87.5\% sensitivity, 88\% specificity), while Yoo et al. used pre-trained CNNs (ResNet50, Inception-v3, MobileNet-v2), with MobileNet-v2 being optimal for smartphones, achieving 95.3\% accuracy (97\% sensitivity, 94.2\% specificity).

Additionally, a mobile app for early Tonsillitis detection processes oral cavity images~\citep{Tonsilitis-detection-app}. Ban et al.~\citep{ban2017detection} demonstrated a smartphone-attachable infrared camera for detecting peritonsillar abscesses in tonsillitis patients, showcasing non-invasive thermography as a viable alternative to CT scans.

\subsection{Others}
\subsubsection{Blood Pressure}
Luo et al.~\citep{luo2019smartphone} proposed a novel contactless method for measuring blood pressure using facial videos captured with a smartphone camera, known as Transdermal Optical Imaging (TOI). TOI separates video frames into RGB bitplanes, extracting hemoglobin-rich signals while discarding melanin-rich ones through an ML-based algorithm. These signals are then merged to generate videos visualizing hemoglobin concentration changes.

A total of 1,328 subjects aged 18 or older, with average reference systolic BP between 100–139 mmHg and diastolic BP between 60–89 mmHg, were included in the study. Signal processing identified 17 ROIs, from which 126 facial transdermal blood flow signals were extracted. PCA was applied for dimensionality reduction, and the features were fed into a multi-layer perceptron to predict systolic BP, diastolic BP, and pulse pressure. The model achieved prediction accuracies of 94.81\% (systolic BP), 95.71\% (diastolic BP), and 95.75\% (pulse pressure). Hypertensive and hypotensive subjects were excluded from the study scope.



\subsubsection{Jaundice}
Jaundice can be detected from smartphone-captured images of a person’s eye by analyzing the sclera (the white portion of the eye) using computer vision techniques. Zhou et al~\citep{zhou2011new} proposed an RGB and HSV color space masking technique for sclera segmentation, while Mariakakis et al.~\citep{mariakakis2017biliscreen} utilized the GrabCut algorithm. Using a dataset of 56 images, Mariakakis et al. achieved 95.7\% sensitivity and 97.4\% specificity in jaundice detection with a Random Forest classifier.


\subsubsection{Parkinson's Disease (PD)}
n video-based analysis of Parkinson's Disease (PD), the initial step involves selecting frames from the video. Ali et al.~\citep{ali2020spatio} utilized a CNN-based classifier to identify the relevant frames. For the final stage, a CNN with a VGG16 architecture~\citep{simonyan2014very} is employed to detect important hand-motor movements from the selected frames. The tasks are assessed separately for two-class classification (PD vs. non-PD) or three-class classification (non-PD, PD with medication, PD without medication). The results show that the two-class classification, with approximately 80\% accuracy, 80\% precision, and 88\% recall, is more reliable than the three-class classification, which achieves 60\% accuracy, 65\% precision, and 60\% recall.
\subsubsection{Post-traumatic Stress Disorder (PTSD)}
Farhana et al.~\citep{shahid2020leveraging} used free-hand sketch images from 160 individuals to diagnose PTSD based on three features: the number of corners, the number of strokes, and the average stroke length. Using only sketching data and the gender of the individuals, their model achieved an accuracy of 82.9–87.8\%. When EEG data was combined with sketching data, the model's accuracy improved significantly to 99.29\%.

\subsubsection{Mental Health Diseases}
Zunino et al.~\citep{Zunino8545095} proposed a solution for early autism diagnosis using video sequences, leveraging differences in simple motor actions between pathological and healthy subjects. Similarly, Tian et al.~\citep{Tian8784802} developed a One Glimpse Autism Spectrum Disorder Detection (O-GAD) network, focusing on repetitive behavioral analysis from video streams.

\subsubsection{Gait Abnormalities}
Aroojis et al.~\citep{aroojisusability} utilized a handheld smartphone with a slow-motion video feature to measure the Edinburgh Visual Gait Score (EVGS) in children with cerebral palsy. Videos of the coronal and sagittal planes were recorded as patients walked, and specific angles were measured by pausing the slow-motion video and using a motion analysis app on the smartphone. Two Pediatric Orthopedic Fellows independently computed the EVGS, achieving complete agreement rates between 61.7\% and 92.5\%.. 

\subsubsection{Dehydration}
In~\citep{saha2022dehydration}, the authors developed a smartphone-based application for dehydration detection using facial image features. The app captures a facial image, calculates dehydration scores for multiple facial landmarks using a Siamese network-based ensemble learning algorithm, and computes a weighted average of these scores to classify whether the individual is dehydrated. Since no public dataset was available, the authors created their own by collecting paired hydrated and dehydrated facial images through the app's guided interface.